\documentclass[11pt]{article}

\usepackage[final]{acl}

\usepackage{times}
\usepackage{latexsym}
\usepackage{multirow}

\usepackage[T1]{fontenc}

\usepackage[utf8]{inputenc}

\usepackage{microtype}

\usepackage{inconsolata}

\usepackage{graphicx}
\usepackage{booktabs}

\usepackage[dvipsnames]{xcolor}
\usepackage[table]{xcolor}

\title{When Misinformation Speaks and Converses: \\Rethinking Fact-Checking in Audio Platforms}

\author{
Chaewan Chun$^{1}$ \and
Delvin Ce Zhang$^{2}$ \and
Dongwon Lee$^{1}$ \\
$^{1}$The Pennsylvania State University, USA \\
$^{2}$University of Sheffield, UK \\
\texttt{czc5884@psu.edu, delvin.ce.zhang@sheffield.ac.uk, dongwon@psu.edu}
}

\begin{document}
\maketitle

\begin{abstract}

Audio platforms have evolved beyond entertainment. They have become central to public discourse, from podcasts and radio to WhatsApp voice notes and live streams. With millions of shows and hundreds of millions of listeners, audio platforms are now a major channel for misinformation. Yet existing fact-checking pipelines are mostly designed for written claims, overlooking the unique properties of spoken media. We argue that audio misinformation is not merely textual content with transcripts: it is structurally different because it is both \emph{spoken}—carrying persuasive force through prosody, pacing, and emotion—and \emph{conversational}—unfolding across turns, speakers, and episodes. These dual properties introduce verification difficulties that traditional methods rarely face. This \emph{position paper} synthesizes evidence across modalities and platforms, examines datasets and methods, and highlights why existing pipelines fail on audio. We argue that advancing fact-checking requires rethinking verification pipelines around the spoken and conversational realities of audio.

\end{abstract}

\section{Introduction}

Spoken media have become a dominant channel for news and commentary. From WhatsApp voice notes and live-streamed talk shows to podcasts, audio platforms now command hours of daily attention, shaping public discourse at scale. Podcasts alone now exceed 4.3 million distinct shows, reaching an estimated 500 million listeners globally, with average consumption of about seven hours per listener each week.\footnote{\url{https://www.demandsage.com/podcast-statistics/}} Podcast listenership has grown steadily year over year, reflecting how audio has shifted from a niche medium to a mainstream source of information and opinion.\footnote{\url{https://podcastatistics.com/}} Yet while audience adoption accelerates, research on fact-checking methods has not kept pace. Automated fact-checking remains text-centric \cite{158}, leaving the spoken and conversational dimensions of misinformation 
underexplored.

\begin{figure}[!t]
  \centering
  \hspace{-0.3in}\includegraphics[width=1.07\columnwidth]{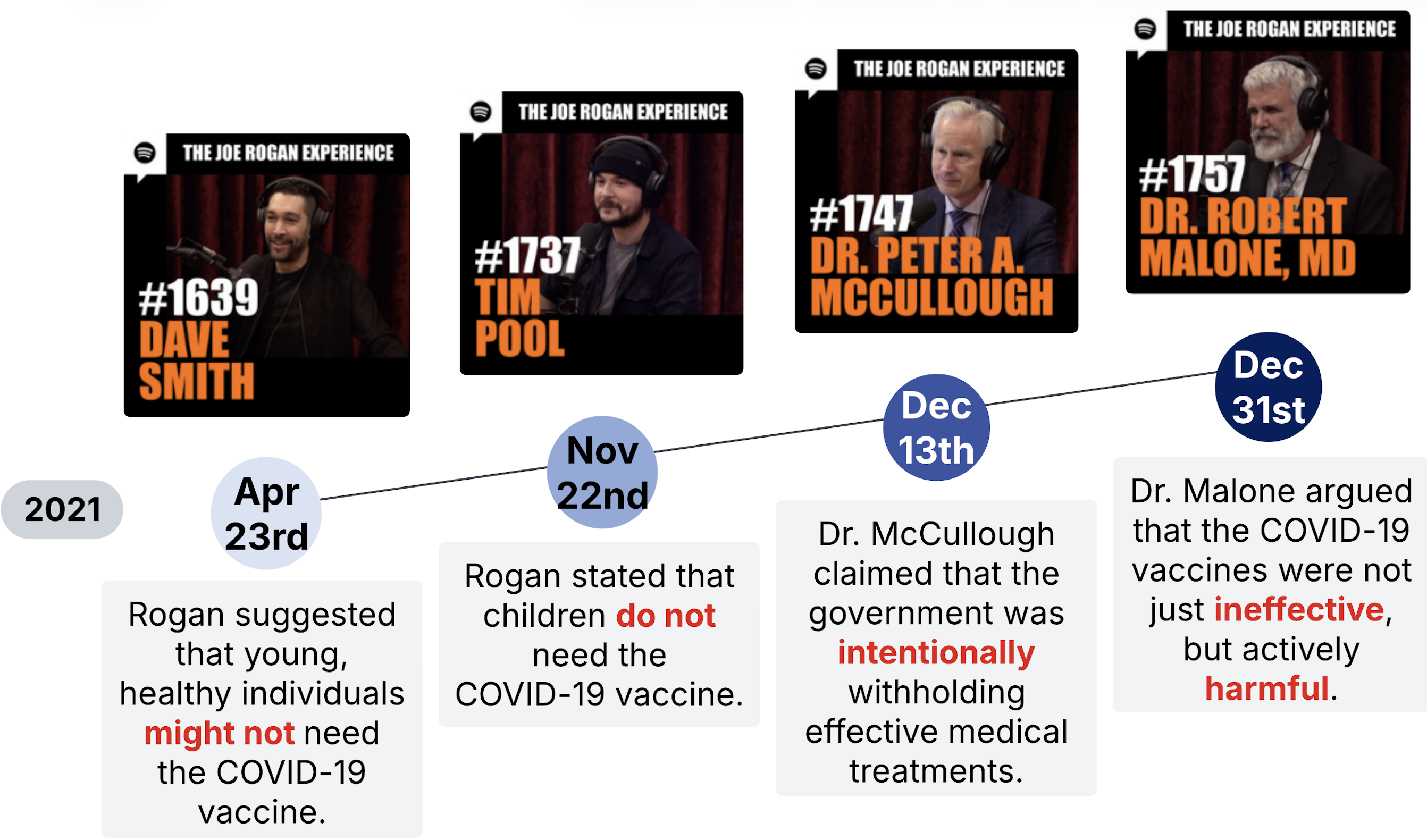}
  \caption{Timeline of cross-episode misinformation on COVID-19. These episodes illustrate how persistent and unchallenged misinformation can accumulate over time. Red text highlights false claims in audio content. Image credit: Spotify}
  \label{fig:ill4}
\end{figure}

The consequences are clear: when false claims circulate in podcasts, radio shows, or private group chats, they spread without the anchoring mechanisms available in text (hyperlinks, citations, overlays). Delivery through tone and pacing lends credibility \cite{guyer2021}, while repetition across turns and episodes reinforces false narratives (see Figure~\ref{fig:ill4}). A single falsehood voiced by a familiar speaker can gain persistence and authority that written text rarely achieves.

What makes audio platforms distinctive and challenging is that they are both {\em spoken} and {\em conversational}. As spoken media, they carry persuasive force through prosody, pacing, and emotion, often making content sound compelling regardless of veracity. As conversational media, they distribute meaning across dialogue history, with context-dependent reference, stance shifts, and cumulative repetition that complicate verification. These dual properties introduce verification demands absent from other modalities, underscoring why audio cannot be treated as \emph{just another transcript}.

We focus on content-level fact-checking in spoken, conversational audio: given what is said, we ask how systems should detect claims, find relevant evidence, decide whether those claims are true, and explain their decisions. We do not address media-integrity questions such as whether an audio signal or voice is synthetic or cloned; those are the target of deepfake detection methods. Our contribution is to show how fact-checking systems themselves must be redesigned for long-form, dialogic speech—where meaning is shaped by delivery, turn-taking, and interaction—regardless of whether the audio was produced by a human or a model.

Fact-checking began as a language-centered problem: systems identified claims in text, retrieved evidence from encyclopedic sources, assessed veracity, and increasingly produced textual rationales \cite{104,105,103,161}. With audio platforms now mainstream, we extend the established approach to spoken dialogue. We (1) show why audio platforms matter and how they differ from other media, (2) survey datasets spanning audio, dialogue, and factuality, highlighting their coverage and limitations, and (3) analyze why existing pipelines fall short and review emerging efforts toward audio-based fact-checking. Taken together, these contributions aim to reposition audio platforms not as a blind spot but as the next frontier for fact-checking research.


\begin{table}[t]
\centering
\small
\begin{tabular}{p{0.45\columnwidth} p{0.4\columnwidth}}
\hline
\textbf{Spoken} & \textbf{Conversational} \\
\hline
Prosody \& emotion        & Adjacency \& repairs \\
Continuous signal         & Distributed claims \\
Incidental sounds \& overlap & Chronology \& callbacks \\
Serialized in-flow listening      & Narrative episodes \\
Acoustic persuasion \& detection   & Parasocial trust \\
\hline
\end{tabular}
\caption{Spoken and conversational properties of audio misinformation and the distinct challenges they introduce for fact-checking.}
\label{tab:spoken_conversational}
\end{table}

\section{Why Audio Platforms Matter}
\label{sec2}

Audio platforms reshape how misinformation spreads and how fact-checking must operate. Unlike standalone text, spoken dialogue carries meaning in delivery and interaction. As a \emph{spoken} medium, prosody, pacing, and vocal cues alter perception; as a \emph{conversational} medium, meaning unfolds across turns, speakers, and episodes. These properties simultaneously amplify persuasion and complicate verification, requiring methods that listen to delivery and track dialogue (Table~\ref{tab:spoken_conversational}).

\subsection{Spoken Misinformation} 

Audio persuades differently from text. Even with identical words, prosody (pitch, intensity, timing), pacing, and emotion shift judgments of confidence, credibility, and intent. Listeners infer trustworthiness from vocal delivery \cite{guyer2021}; deception studies show that speech rate, intensity, pitch variation, and hesitations influence truthfulness judgments \cite{18,22}; and even eyewitness testimony accuracy has been linked to vocal cues, indicating that delivery can act as a heuristic shortcut \cite{gustafsson}. At scale, analyses of 88{,}000 podcast episodes show that vocal qualities (e.g., energy, seriousness) predict engagement, with adversarial acoustic features outperforming common baselines \cite{41}. These persuasion effects interact with interpretation. Irony perception depends on prosody; without auditory cues, sarcasm is often taken literally \cite{matsui, surveysarcasm, zhuli2025}. On audio platforms, sarcastic and ironic intent relies on delivery \cite{rao2022,li2025}, which can tilt credibility judgments and downstream decisions. Together, these results place prosodic and other paralinguistic features at the center of how people absorb, interpret, and evaluate information.

Meaning in audio unfolds as a continuous signal. Claims often span prosodic units rather than tidy sentence boundaries. Hesitations, elongations, and timing cues can split the relevant proposition across adjacent acoustic spans, even within a single speaker's turn. Speakers slow down and insert disfluencies before more surprising words, effectively ‘buying time,’ and speech rate adapts to information content \cite{umyeah}. Podcast discourse further complicates scope: episodes are long, digressive, and conversational, with ads and side-tracks that introduce long-range dependencies and noisy content—making chapterization/segmentation a prerequisite for downstream tasks like retrieval \cite{narrative2024,podtile}. 

Incidental sounds (laughter, music, effects) and overlapped speech (interruptions, backchannels, crosstalk) can obscure boundaries and degrade segmentation/diarization, motivating explicit overlap detection \cite{overlap2021,overlap2025}. Serialized listening sustains narratives over time; surveys report about 80\% of listeners finish most episodes they start \cite{37}. Case studies show the same dynamics across platforms: WhatsApp voice notes in Portugal circulated urgent COVID-19 rumors and policy critiques \cite{6}, while an analysis of leading U.S. political podcasts documented frequent false or unsubstantiated claims reaching large audiences \cite{audioReckoning}. Unlike text feeds where visual labels sit on posts, podcasts and streams are consumed in-flow, so the persuasive payload reaches listeners during listening itself \cite{10}.

The same cues that make audio persuasive also enable detection. Because voices carry stance and identity cues—intonation, stress, disfluencies—acoustic evidence can supplement or even surpass transcripts for certain phenomena. In political debates, adding acoustic features significantly improves identification of check-worthy claims, and in some conditions audio-only models outperform text-only systems \cite{21}. Multimodal sarcasm corpora likewise find prosody outperforms text for irony and emotion \cite{li2025}. More broadly, multimodal misinformation frameworks find that acoustic encoders capture semantic-level cues that transcripts miss, yielding gains when fused with other modalities \cite{5}; and while standalone deception cues can be noisy, features like jitter, pauses, and pitch instability become informative when integrated with lexical and visual signals \cite{18,22}. These observations motivate audio-aware fact-checking that listens to delivery and sequence, not only to words.

\subsection{Conversational Misinformation}

Conversation changes what a “claim” is. In dialogue, propositions are negotiated through adjacency, uptake, repairs, hedges, and challenges; a first mention is rarely the final form. Claims are also distributed and compositional—parts contributed at different turns and by different speakers that only add up to a verifiable unit when considered together \cite{specialized_models,deng2024}. Chronology further shapes judgment: primacy and recency biases anchor interpretation, and repeated callbacks increase availability and felt plausibility \cite{35}; podcast episodes often include narrative content and recurring formats, which can sustain storylines over time \cite{narrative2024}. Social roles and bonds intensify these effects: hosts and recurring guests shape expectations and perceived credibility; repeated agreement can signal consensus, and parasocial trust makes familiar voices feel credible \cite{oneofus,nostalgia}. In short, conversational structure itself—how things are said, when they are said, and by whom—drives persuasion independent of the words alone.

Misinformation leverages these dynamics. On messaging platforms, misleading WhatsApp voice notes follow repeatable rhetorical templates—greeting, insider/expert positioning, emotional appeals, a central false assertion, and a call-to-action \cite{6,7}. Long-form podcasts exploit repetition and serialized exposure: narratives are seeded, revisited, and strengthened across episodes. For example, \emph{The Joe Rogan Experience}, one of the world’s most influential podcasts (Figure~\ref{fig:ill4}), advanced false COVID-19 vaccine narratives across multiple episodes. In April, he suggested young, healthy people might not need vaccinations \cite{152}; by November, he argued against vaccinating children \cite{151}; and in December, with guests Dr. Peter A. McCullough and Dr. Robert Malone, he claimed vaccines were ineffective or harmful \cite{149,150}. This cross-episode repetition amplified misinformation through cumulative exposure. In January 2022, over 270 U.S. health professionals issued an open letter urging platform action \cite{156}. These dynamics scale beyond a single program. U.S. podcast studies describe “toxic conversation chains”—emotionally charged exchanges that sustain harmful narratives within episodes—and document that such toxicity recurs across many episodes \cite{31}.

These dynamics complicate verification as well as persuasion. Empirically, systems perform markedly better when given dialogue context rather than isolated utterances \cite{mad2}, and human annotators likewise use surrounding context to judge check-worthiness \cite{27}. To make this more concrete, we draw on MAD2, a benchmark of 1{,}000 two-speaker English dialogues (about 10 hours of audio) with 8{,}192 sentences and 3{,}368 check-worthy claims, each annotated with a binary true/false label \cite{mad2}. Each dialogue is accompanied by a full transcript, and every claim is aligned to its spoken span using word-level timestamps, so models can operate over precise seconds-based context windows rather than only sentence IDs. The dialogues are multi-turn, speaker-attributed sequences, and in our analysis, we consider two simple context regimes for a standard text-only verifier: (i) only the 15 seconds preceding the claim, and (ii) the full dialogue surrounding the claim.

To isolate the role of order, we compare a base model to a shuffled-turns variant that randomizes the non-claim utterances within each dialogue while keeping the lexical content fixed on MAD2. For each context regime (past 15 seconds vs.\ full dialogue), Table~\ref{tab:tags_shuffle} reports F1 and AUC for the base model and its shuffled counterpart. Across both regimes, shuffling consistently reduces performance despite identical words, indicating that temporal order and pacing carry predictive signal beyond lexical content. This small, illustrative experiment complements prior findings that speech rate and timing track information structure \cite{umyeah}, and supports our broader claim that conversational order and accumulation shape what both verifiers and annotators perceive as check-worthy and true.


\begin{table}[t]
\centering
\small
\setlength{\tabcolsep}{4pt}
\renewcommand{\arraystretch}{1.1}
\begin{tabular}{l|l|ccc}
\hline
\textbf{Metric} & \textbf{Setting}  & \textbf{Past 15 sec} & \textbf{Full dialogue} \\
\hline
\multirow{2}{*}{F1}   & Base  & 0.6799 & 0.7140 \\  
     & Shuffled  & 0.6590 & 0.7041 \\
\hline
\multirow{2}{*}{AUC}  & Base & 0.7408 & 0.7924 \\  
  & Shuffled  & 0.7203 & 0.7780 \\
\hline
\end{tabular}
\caption{
Illustrative ablation on MAD2 \cite{mad2}: F1 and AUC for text-only claim verification under two context regimes. Rows compare a \textit{Base} model that reads turns in chronological order versus a \textit{Shuffled} variant that randomizes non-claim turn order while keeping content fixed. Shuffling reduces performance in both settings, implying that chronological turn order carries predictive signal beyond lexical content.
}

\label{tab:tags_shuffle}
\end{table}


\begin{table*}[!t]
\centering
\scriptsize
\setlength{\tabcolsep}{3pt}
\renewcommand{\arraystretch}{1.15}
\resizebox{\textwidth}{!}{%
\begin{tabular}{l|c|c|c|c|c|c|c}
\toprule
\textbf{Dataset} & \textbf{Audio} & \textbf{Dialogue} & \textbf{Transcript} & \textbf{Factuality} & \textbf{Size} & \textbf{Lang} & \textbf{Domain} \\
\hline
LibriSpeech \cite{librispeech} & x & -- & x & -- & 1k hrs & En & audiobooks \\
VoxPopuli \cite{voxpopuli} & x & -- & x & -- & 400k hrs raw / 19.1k hrs labeled & Multi & parliamentary speeches \\
People's Speech \cite{peoplespeech} & x & -- & x & -- & 30k+ hrs & En & diverse speech \\
LibriHeavy \cite{libriheavy} & x & -- & x & -- & 50k hrs & En & audiobooks \\
SSSD \cite{sssd} & x & x & -- & -- & 700+ hrs & En & everyday conv. \\
\hline
Ubuntu Corpus \cite{ubuntucorpus} & -- & x & -- & -- & 930k dialogs & En & tech support \\
DailyDialog \cite{dailydialog} & -- & x & -- & -- & 13.1k dialogs & En & daily chat \\
Persona-Chat \cite{personachat} & -- & x & -- & -- & 11k dialogs / 164k utt. & En & open-domain, persona \\
MultiWOZ \cite{multiwoz} & -- & x & -- & -- & 10k dialogs & En & task-oriented, multi-domain \\
GroundedConv \cite{groundconv} & -- & x & -- & -- & 4.1k dialogs & En & Wikipedia-grounded movie chat \\
Topical-Chat \cite{topical} & -- & x & -- & -- & 11.3k dialogs & En & knowledge-grounded chat \\
Wizard of Wikipedia \cite{wizardwiki} & -- & x & -- & -- & 22k dialogs / 202k utt. & En & knowledge-grounded chat \\
SAMSum \cite{samsum} & -- & x & -- & -- & 16.4k dialogs & En & messenger chat, abstractive summarization \\
EmpatheticDialogues \cite{empatheticDialog} & -- & x & -- & -- & 25k dialogs & En & emotion-grounded chat \\
BlendedSkillTalk \cite{put2020} & -- & x & -- & -- & 6.8k dialogs & En & open-domain, blended skills \\
MultiWOZ 2.1 \cite{multiwoz21} & -- & x & -- & -- & 10k dialogs & En & task-oriented, multi-domain \\
MedDialog \cite{meddialog} & -- & x & -- & -- & 0.26m / 3.4m dialogs & En/Zh & medical \\
WhatsApp 2021 \cite{whatsapp21} & -- & x & -- & -- & 298k msgs & Pt & political WhatsApp groups, link/media typology \\
MediaSum \cite{mediasum} & -- & x & x & -- & 463.6k dialogs, summaries & En & interviews (NPR/CNN) \\
CMCC \cite{cmcc} & -- & x & x & -- & 100k dialogs (8.9k labeled) & Zh & customer service \\
HANSEN \cite{hansen} & -- & x & x & -- & 17 datasets / $\sim$21k AI samples & En & spoken authorship \\
Audio Dialogues \cite{audiodialogues} & -- & x & -- & -- & 163.8k dialogs & En & audio/music understanding \\
Liu2025 Bilingual Dialogue \cite{liu2025biling} & -- & x & -- & -- & see repo & En/Zh & personality, emotion \\
\hline
MELD \cite{meld} & x & x & x & -- & 1.4k dialogs / 13k utt. & En & TV show (Friends), emotion recognition \\
AVSD \cite{avsd} & x & x & x & -- & 11.8k dialogs / 118k QA pairs & En & video-grounded daily activities \\
Spotify Podcasts \cite{spotify} & x & x & x & -- & 60k hrs / 100k+ eps & En & podcasts \\
CHiME-6 \cite{chime6} & x & x & x & -- & 40+ hrs & En & dinner-party conversations, conversational ASR \\
DiPCo \cite{dipco} & x & x & x & -- & 5.3 hrs / 10 sessions & En & dinner-party conversations \\
SPGISpeech \cite{spgispeech} & x & x & x & -- & 5k+ hrs & En & earnings calls \\
Earnings-21 \cite{earnings21} & x & x & x & -- & 39 hrs & En & earnings calls \\
MD3 \cite{md3} & x & x & x & -- & 20 hrs & En & information-sharing tasks \\
CANDOR \cite{candor} & x & x & x & -- & 850 hrs / 1,656 dialogs & En & everyday conv. (video chat) \\
DailyTalk \cite{dailytalk} & x & x & x & -- & 20 hrs / 2,541 dialogs & En & conversational TTS \\
SPoRC \cite{sporc} & x & x & x & -- & 1.1m eps & En & podcasts \\
MultiDialog \cite{multidialog} & x & x & x & -- & 340 hrs / 9k dialogs & En & open-domain, audiovisual \\
SPGISpeech2 \cite{spgispeech2} & x & x & x & -- & 3.78k hrs & En & earnings calls \\
CASPER \cite{casper} & x & x & x & -- & 3 hrs & En & spontaneous conv. \\
DeepDialog \cite{deepdialog} & x & x & x & -- & 488 hrs / 40.2k dialogs & En & 41 domains, 20 emotions \\
\hline
PHEME \cite{pheme} & -- & x & -- & * & 1,185 threads & En & Twitter rumors \\
ClaimBuster \cite{78} & -- & x & x & * & 23k sentences & En & political debates \\ 
Audio Check-Worthiness \cite{21} & x & x & x & * & 48 hrs / 34.5k sents & En & political debates, speeches, interviews \\
ViClaim \cite{173} & -- & -- & x & * & 1.8k videos / 17.1k sents & En/De/Es & YouTube short videos, claim detection \\
CT-FCC-18 \cite{ctfcc18} & x & -- & x & x & 33 min / 286 claims & En & political debates \\
WhatsApp 2019 \cite{whatsapp19} & -- & x & -- & x & 912k msgs & Pt & WhatsApp political groups \\
CI-ToD \cite{citod} & -- & x & -- & x & 3,190 dialogs & En & task-oriented, consistency labels (HI/QI/KBI) \\
DialFact \cite{dialfact} & -- & x & -- & x & 22.2k claims & En & fact-checking in dialogue \\
\hline
Fact-Checking Podcasts \cite{setty2025} & x & x & x & x & 531 eps / ~2.0k utt. (annot.) & En/No/De & podcasts (news, health) \\
\rowcolor{green!20}
\textbf{MAD} \cite{mad1} & x & x & x & x & 600 dialogs / 4.9k sents & En & spoken dialogue \\ 
\rowcolor{green!20}
\textbf{MAD2} \cite{mad2} & x & x & x & x & 1k dialogs / 8.2k sents / word-level ts & En & spoken dialogue \\
\bottomrule
\end{tabular}
}
\caption{Representative datasets across audio, dialogue, transcript, and factuality dimensions. \\
$^{*}$: check-worthiness only (no veracity labels). \\
\textit{Abbrev.:} QA = question answering; TTS = text-to-speech; ts = timestamps.
}
\label{tab:unified_datasets}
\end{table*}

\section{Datasets}

Research into fact-checking for audio platforms remains relatively nascent. As summarized in Table~\ref{tab:unified_datasets}, most corpora address only subsets of these dimensions and rarely combine all four dimensions—audio, dialogue, transcripts, and factuality.

Large-scale speech corpora exist but target automatic speech recognition (ASR) or representation learning, not verification. Read audiobooks (LibriSpeech \cite{librispeech}, LibriHeavy \cite{libriheavy}), multilingual parliamentary speeches (VoxPopuli \cite{voxpopuli}), and English ASR collections (People’s Speech \cite{peoplespeech}) offer thousands to hundreds of thousands of hours, yet lack claim units, timestamped rationales, and veracity labels. These corpora improve speech modeling, not veracity evaluation.

Dialogue benchmarks have expanded but remain text-only. Early datasets (Ubuntu \cite{ubuntucorpus}, DailyDialog \cite{dailydialog}, Persona-Chat \cite{personachat}) advanced turn-taking, persona, and emotion, while later work emphasized empathy, knowledge grounding, or summarization (e.g., Topical-Chat \cite{topical}, Wizard of Wikipedia \cite{wizardwiki}, SAMSum \cite{samsum}). These corpora are invaluable for conversational modeling, but they lack audio and do not provide veracity labels.

A separate line of work scales spoken dialogue without factuality. These corpora combine audio, diarization, and transcripts—supporting prosody, overlap, and long-context modeling—but they do not label whether claims are true. Representative examples span emotion/video-grounded conversations (MELD \cite{meld}; AVSD \cite{avsd}), multiparty household talk captured with distant microphones (CHiME-6 \cite{chime6}), and multi-speaker earnings-call speech (SPGISpeech \cite{spgispeech}); large long-form conversational releases broaden duration and domains (e.g., Spotify Podcasts \cite{spotify}; CANDOR \cite{candor}). Together, they are essential for modeling how something is said, not for assessing whether it is true.

Where factuality is present, coverage is often partial or limited to a single modality.
Some datasets provide check-worthiness only—flagging what to fact-check but not whether it is true (PHEME \cite{pheme}; ClaimBuster \cite{78}; ViClaim \cite{173}). Others include veracity but miss key modalities: CT-FCC-18 \cite{ctfcc18} aligns short debate audio to fact-checked claims without full dialogue context, while DialFact \cite{dialfact} verifies claims in textual dialogues with no audio. Related work benchmarks consistency rather than truth \cite{citod}, and WhatsApp 2019 \cite{whatsapp19} adds veracity assessments to media shared in chats but does not capture audio. Across these resources, the common gaps include variable claim granularity, limited timestamping, narrow domains, and weak multi-evidence support.

Only recently have datasets appeared that cover audio, dialogue, transcripts, and veracity. \citet{setty2025} targets podcasts with transcripts plus check-worthiness and supports/refutes annotations. MAD \cite{mad1} introduces multi-turn spoken dialogues with aligned audio and veracity labels, and MAD2 \cite{mad2} provides roughly 1,000 dialogues with thousands of check-worthy claims and word-level timestamps. 
Despite progress, current resources remain small-scale, English-dominant, and narrow in domain coverage—underscoring the urgent need for large-scale, multimodal audio fact-checking datasets.

\section{Where Traditional Pipelines Fail}
\label{sec:pipeline}

Most fact-checking systems follow a four-stage pipeline: Claim Detection (CD) identifies check-worthy statements; Evidence Retrieval (ER) queries trusted sources to gather passages relevant to the claim; Claim Verification (VER) compares the claim against the retrieved evidence to assign a verdict (supported, contradicted, or insufficient) and a confidence score; and Explanation Generation (GEN) produces a human-readable rationale that highlights the evidence and explains the verdict. Building on our earlier discussion of how audio platforms differ, we now examine where this pipeline breaks for spoken dialogue. Table~\ref{tab:audio_pipeline} summarizes typical models, their failure modes on spoken dialogue, and the design requirements we argue for at each stage of the pipeline.

\subsection{Claim Detection (CD) Task}

Traditional CD treats one sentence as a single proposition, but conversational audio rarely obliges: claims spread across adjacent turns, carry hedges, or sit inside Q\&A and anecdotes, so single-turn sentence classifiers miss the claim or mis-scope its span \cite{specialized_models,deng2024}. Text-only detectors ignore how delivery moves meaning: sarcasm, emphasis, emotion, and intent live in prosody—pitch, timing, and intensity—so models that ignore nonverbal speech cues misread intent \cite{biron2025,lexical2008}. Familiar hosts, recurring guests, and conversational role structure (e.g., host/guest/caller) shape interaction patterns and perceived salience, so CD should encode who is speaking and how they interact, not just what they say \cite{whoisspeaking,stemaaai,sourceclaim}. Treating turns independently discards uptake, repairs, and repetition trajectories that strengthen or revise a proposition over time \cite{whoisspeaking}. Empirically, order carries signal: shuffling turns weakens downstream prediction and, on MAD2, degrades accuracy (Table~\ref{tab:tags_shuffle}) \cite{umyeah,mad2}. Relatedly, argument-aware summarization and key-point analysis aggregate across turns to surface conversation-level claims not stated in any single sentence \cite{convosumm}. 

Beyond local phrasing, conversational claims depend on dialogue history: meaning is often incomplete without preceding turns, and delivery can amplify perceived salience independent of truth. For instance, misleading WhatsApp voice notes often follow a repeatable arc—opening with a personal greeting, asserting source credibility (insider/expert/eyewitness), leveraging negative emotional tone (panic, fear, or anger), delivering the claim, and—in about one-third of cases—urging recipients to forward it \cite{7}. When flattened to text, these cues disappear, and models mistake rhetorical mobilization for factual salience—a recurring source of false positives and missed context.

Upstream artifacts further distort the "claim unit." ASR segmentation can split or merge spans; diarization errors swap speakers; and overlapped talk stresses voice activity detection (VAD) and segmentation, where thresholding can clip turn onsets/offsets - each of which destabilizes CD in long-form talk \cite{asr2021,overlap2021, whisperx, overlap2025}. Real-world conditions magnify the problem: Whisper reports $\sim$2.5-2.7\% word error rate (WER) on LibriSpeech test-clean \cite{radford}, yet podcast corpora reach $\sim$18.1\% WER and include non-speech/extraneous segments that must be filtered \cite{spotify,narrative2024}. Earlier end-to-end ASR and short-window decoders degrade under noise; in contrast, large-scale pretraining, timestamped decoding/VAD (e.g., Whisper), and forced alignment (WhisperX) improve robustness and alignment in noisy or low-resource settings \cite{radford,whisperx}. Dataset mirrors this: sentence-level check-worthiness often fails under ellipsis but improves with decontextualization \cite{specialized_models}; audio/disfluency cues can match or beat text-only baselines in multi-speaker setups (HuBERT \cite{hubert}, wav2vec~2.0 \cite{wav2vec2}) \cite{21}; yet speech-native CD datasets remain small \cite{21}.

Methodologically, CD research has moved from early feature-based systems \cite{78,79} to neural rankers and pre-trained language models such as BERT and RoBERTa \cite{bert,roberta}, with widespread adoption in recent CD work \cite{192}. Subsequent work explores claim-attribute modeling \cite{76} and impact-aware prioritization \cite{192}. Yet most work remains source- and modality-specific \cite{192}.

To close this gap, several strands of work are converging. \emph{Conversation-first datasets} such as DialFact explicitly surface colloquialisms, ellipsis/coreference, and context dependence that challenge sentence-isolated detectors, enabling dialogue-aware baselines and evaluation \cite{dialfact}. \emph{Claim-unit reconstruction} via decontextualization assembles a self-contained proposition before scoring and improves extraction quality and readiness for downstream components \cite{deng2024, fan2025}. \emph{Prosody- and role-aware modeling} incorporates pitch/timing/intensity cues and speaker/interaction structure to better capture communicative function and salience \cite{biron2025,whoisspeaking,stemaaai}. Finally, live CD systems (e.g., LiveFC) bring streaming transcription and online diarization together with windowed claim detection/normalization, so detectors consume temporally ordered, speaker-attributed candidates rather than isolated sentences \cite{28}. In practice for CD: model dialogue acts and local context \cite{whoisspeaking}; encode speaker roles and interaction structure \cite{whoisspeaking,stemaaai}; condition on prosody \cite{lexical2008,biron2025}; and aggregate adjacent turns into a standalone claim unit \cite{deng2024}.

\begin{figure}[tb]
  \centering
  \includegraphics[width=1\columnwidth]{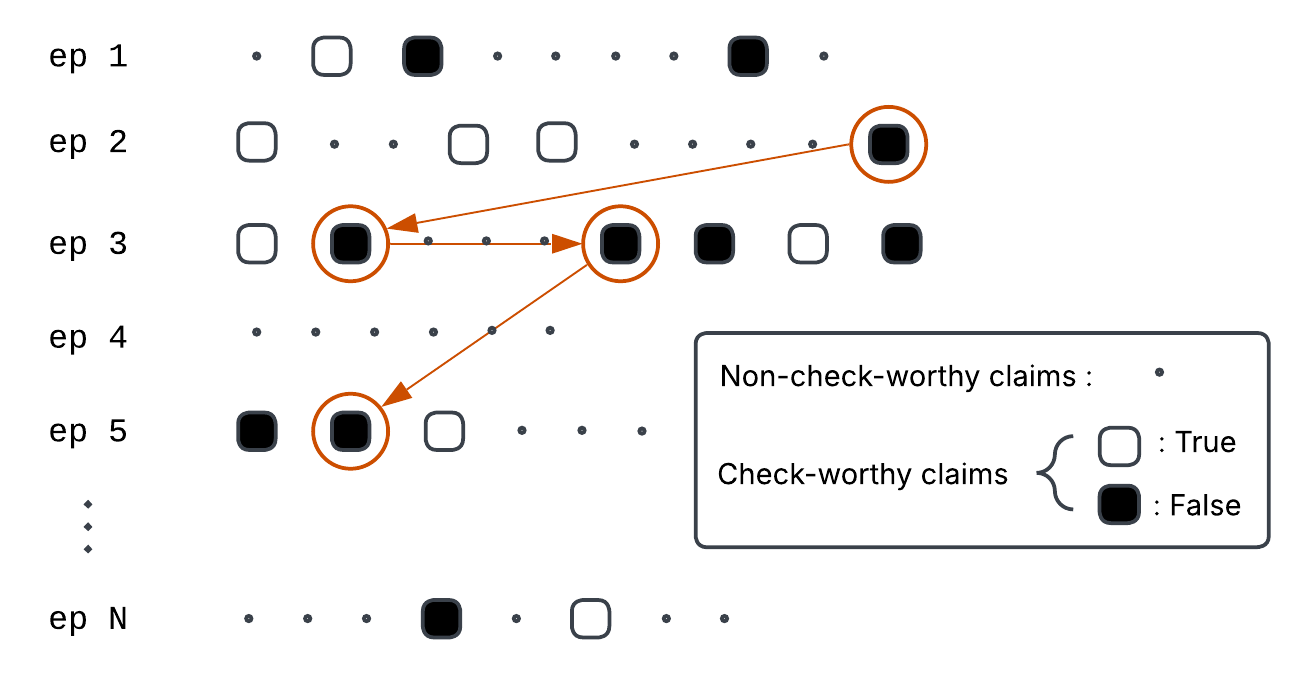}
  \caption{Illustration of claim detection and verification tasks in podcasts.}
  \label{fig:ill-ep}
\end{figure}

\subsection{Evidence Retrieval (ER) Task}
Traditional ER is built for clickable text: models often assume sentence-level evidence, as in FEVER, a standard textual fact-checking benchmark \cite{103}, that lexical retrieval can fetch as one or more sentence spans. Spoken platforms break these assumptions. Audio is continuous and not clickable: meaning is distributed across turns, speakers, and episodes, so evidence must be located in time, not just on a page \cite{podtile}. As a result, single-turn lexical queries miss callbacks, cross-turn references, and off-mic context; colloquialisms, slang, and code-switching change the retrieval error profile \cite{spotify,transcriptnoise}; and pipelines that ignore chapter titles and episode-level metadata (titles/descriptions/RSS IDs (Really Simple Syndication)) leave strong production-metadata constraints unused \cite{podtile,transcriptnoise}. Span-level claim extraction can also improve retrieval relative to full-post queries \cite{twitterspan}. To keep verification auditable, ER therefore has to return time-anchored evidence—an episode identifier plus start/end seconds—so downstream components can attribute who said what, when, and support revisions when on-air corrections appear \cite{podtile}.

In real deployments, upstream artifacts complicate this. ASR segmentation can split or merge spans, diarization uncertainty blurs speaker identity, and overlapped talk stresses VAD/segmentation—each interacts with lexical retrieval and can hide the very moments VER needs \cite{overlap2021,transcriptnoise,overlap2025}. Confidence-only filters are unreliable for detecting ASR errors \cite{asr2025}. Crucially, error types (punctuation, proper nouns \cite{seq2seq}, code-switch points)—not just average WER—drive indexing and alignment errors \cite{transcriptnoise,spotify}. These conditions push ER beyond plain text toward the selective use of acoustic/phonetic cues and tighter use of production metadata.

Resources mirror the gap. Wizard-of-Wikipedia offers turn-level grounding but presumes explicit passages rather than timestamped audio \cite{wizardwiki}; ClaimBuster and DialFact add check-worthiness/verification labels yet are largely text-only \cite{78,dialfact}. Speech-native efforts go further: Audio Check-Worthiness couples speech and claims, and the Fact-Checking Podcast dataset preserves ASR and diarization with podcast-native claims, check-worthiness, and rationales—though all remain small relative to text corpora \cite{21,setty2025}. Together, these efforts point toward retrieval that treats audio as a continuous, produced medium.

Methodologically, a speech-native pipeline adopts: \emph{timestamped spans} (transcript slice + episode ID + start/end seconds), indexed with \emph{dual indices} (lexical + optional acoustic/phonetic), then \emph{re-ranking by temporal proximity, speaker match, and meta-structure} (chapters/show notes/episode metadata) \cite{podtile}, while maintaining \emph{long-horizon memory} so recurring assertions can be linked across episodes. In effect, ER outputs a compact bundle (episode, timestamp, transcript slice, speaker posterior, etc.) that VER can consume without re-resolving timing or speaker identity.

\subsection{Claim Verification (VER) Task}
Verification tuned on text typically frames each claim–passage pair as single-span natural language inference (NLI), but—--as Figure~\ref{fig:ill-ep} illustrates---podcast evidence is time-anchored, multi-turn, and often cross-episode, making FEVER-style single-sentence models brittle in practice \cite{103}. Specifically, multi-turn entailment and temporal qualifiers (“last week”, “earlier in the show”) spill beyond a single span, and VER should support revision when on-air corrections appear. Small timing misalignments and diarization uncertainty further depress multimodal performance unless modeled explicitly \cite{evidence2023,overlap2021,overlap2025}. Empirically, FEVER-tuned models degrade in dialogue; DialFact reports sharp drops from cross-turn references and informality \cite{dialfact}; and CI-ToD surfaces contradictions arising from dialogue history itself \cite{citod}. Throughput and latency compound the challenge: re-ranking thousands of candidates with large NLI models is costly on continuous, claim-dense streams, and on live platforms, timing determines whether interventions land in time \cite{28,realtimetrust}.

In response, VER shifts to: \emph{multi-span, time-aware reasoning} over sets of spans; explicit \emph{who/when constraints} to resolve callbacks, coreference, and temporal qualifiers \cite{exfever,multihop,evidence2023}; \emph{alignment-robust fusion}—late/gated combination of audio encoders with text plus masking of diarization uncertainty—so mild desynchronization does not derail entailment \cite{temporalMisalign,overlap2021,overlap2025,reviewSpeakerdiar}; and \emph{WER-aware training}: mixing oracle and ASR transcripts with audio-only views and adding punctuation/noise perturbations, recognizing that error type—not just average WER—drives downstream failure \cite{evidence2023, transcriptnoise}, with practices that carry over from noisy-transcript analyses \cite{transcriptnoise,asr2021}.

Efficiency-focused designs make deployment feasible. Lightweight verifiers such as MiniCheck approach GPT-4-level verification at $\sim$400$\times$ lower cost \cite{25}, while streaming systems like LiveFC run real-time transcription, online diarization, retrieve timestamped windows, and perform incremental detection/verification online \cite{28}. Complementary probes—Debate-to-Detect and DEFAME—stress debate structure and cross-modal cues in realistic, structured settings \cite{debateToDetect,defame}. Finally, \emph{attribution and recoverability} practices from modular retrieval-augmented generation (RAG)—self-reflective retrieval and recoverability audits—transfer naturally to timestamped audio bundles so cited moments can be re-found and audited end-to-end \cite{selfrag,correctfaith,xing2025}.

\subsection{Explanation Generation (GEN) Task}

GEN fails on audio when it treats explanations as text-only. Without audio, listeners cannot hear how delivery shapes interpretation (sarcasm, emphasis, hesitation). Without timestamps and speaker labels, rationales are hard to audit or replay. Explanations often ignore the conversational path - who challenged whom and when - so they cannot justify mid-stream verdict changes in multi-turn dialogue \cite{dialfact}. Multimodal fact-checking and deception-style probes show that adding non-text modalities improves robustness by capturing delivery-related cues \cite{22}. Meanwhile, the literature increasingly explores retrieval-augmented evidence pipelines and end-to-end systems that jointly produce verdicts and rationales \cite{161,evidence2023}, alongside modern large language model (LLM) prompting frameworks for generation and detection \cite{193}. For audio platforms, those strengths must be \emph{audio-native, timestamped, and dialogue-aware} \cite{realtimetrust,28}.

A practical design is to return evidence you can listen to. Each explanation should pair a readable transcript span with a short, click-to-hear clip (3-10s) and explicit speaker attribution. Optionally flag salient prosodic cues so users can judge delivery, and include a brief decontextualized gloss so the claim remains self-contained for readers without losing its audio anchor \cite{deng2024}. Explanations should present a compact \emph{timeline} - first assertion $\rightarrow$ challenge $\rightarrow$ repair - anchored by episode/turn indices, and bundle the timestamped audio/transcript with any external documents so the cited moment can be re-found and audited \cite{evidence2023,podtile}. Live systems already point the way: LiveFC surfaces speaker-attributed snippets with verdicts in real time, and modular/two-stage pipelines - where lightweight detectors gate heavier generators - can deliver rationales at acceptable latency in streaming contexts \cite{28,twostageinfuse,rationale2023}. \emph{Recoverability-based attribution} (mask the cited span and test whether the system can restore it) stabilizes explanations and transfers naturally to timestamped audio bundles \cite{xing2025}. In short, when GEN embraces audio-native affordances and recoverability, it complements timestamped retrieval and time-aware verification rather than inheriting the blind spots of text-only explanations.

\begin{table*}[t]
\centering
\small
\setlength{\tabcolsep}{4pt}
\begin{tabular}{p{0.04\textwidth} p{0.2\textwidth} p{0.33\textwidth} p{0.37\textwidth}}
\hline
\textbf{Task} &
\textbf{Representative Models} &
\textbf{Limitations on Spoken Dialogue} &
\textbf{Proposed Design} \\
\hline
\multirow[c]{5}{=}{\textbf{CD}} &
Sentence-level check-worthiness and claim classifiers over text (BERT-/RoBERTa-style rankers). &
Assume one sentence $\approx$ one claim; miss claims spread across turns, Q\&A, and anecdotes; ignore prosody and speaker roles; brittle under ASR/diarization errors and high WER in podcasts. &
Turn- and speaker-aware detectors that aggregate adjacent turns into decontextualized claim units, condition on prosody, and operate over ordered, speaker-attributed windows (including streaming). \\
\hline
\multirow[c]{5}{=}{\textbf{ER}} &
Lexical retrievers over sentence spans in text corpora. &
Built for clickable text spans, not continuous audio: single-turn queries miss callbacks and cross-episode references; ignore production metadata; ASR errors and code-switching corrupt indices and timing. &
Return time-anchored evidence (episode ID + start/end seconds + transcript slice), indexed with dual lexical/phonetic views and re-ranked using speaker and production metadata, with memory for cross-episode recurrence. \\
\hline
\multirow[c]{6}{=}{\textbf{VER}} &
Single-span claim--passage NLI verifiers, lightweight verifiers, and early multimodal audio--text verifiers. &
Assume single-span, clean text; struggle with multi-turn, time-qualified, and cross-episode evidence; sensitive to timing and diarization errors; FEVER-tuned models degrade on informal dialogue; large verifiers are costly for continuous/live streams. &
Multi-span, time-aware reasoning with explicit who/when constraints; alignment-robust audio--text fusion; WER-aware training (oracle+ASR, noise/punctuation perturbations); cascaded verifiers for streaming, plus attribution and recoverability over timestamped bundles. \\
\hline
\multirow[c]{6}{=}{\textbf{GEN}} &
Text-only rationale generators and retrieval-augmented pipelines that generate explanations jointly with verdicts, often via LLM prompting. &
Treat explanations as pure text: ignore delivery (sarcasm, emphasis, hesitation) and conversational path; lack timestamps and speaker labels; make mid-stream verdict changes hard to justify and leave the link to the underlying audio opaque. &
Timestamped, speaker-attributed explanations that pair short audio clips with readable transcript spans and a compact assertion$\rightarrow$challenge$\rightarrow$repair timeline, using recoverability-based attribution and lightweight two-stage generators suitable for streaming. \\
\hline
\end{tabular}
\caption{Summary of how traditional pipeline components behave on spoken dialogue and the corresponding design needs. See Section~\ref{sec:pipeline} for references.}
\label{tab:audio_pipeline}
\end{table*}

\section{Future Directions}

Moving the field forward requires a few pieces to work together. First come the foundations: build larger speech-native corpora with factuality or check-worthiness labels, aligned transcripts, and speaker/role metadata; augment them with targeted synthetic variants (accents, overlap, punctuation/noise). Then make the pipeline live: pair streaming ASR and online diarization with evidence retrieval that returns time-anchored spans (episode ID plus start/end seconds) via dual lexical–acoustic indices and windowed retrieval; verify with lightweight models that can keep up in live settings. Richer reasoning and accountability follow: use alignment-robust (late/gated) fusion that masks diarization uncertainty so prosody and timing inform decisions; model the conversational fabric—roles, hedges, stance, and uptake/repairs—to assemble the claim unit across turns; track narratives across episodes with simple graph or archival tools; and surface listenable evidence (a transcript snippet plus a short audio clip with speaker attribution), validated by recoverability tests. Finally, make the system trustworthy and equitable: replace confidence-only gates with calibrated, WER-aware verification that is sensitive to error types (punctuation, proper nouns, code-switch points), and mitigate accent/dialect and noise brittleness through dialect-aware pretraining, domain adaptation, and perturbation-based checks.

\section{Conclusion}
Audio platforms reshape misinformation by combining the persuasive force of spoken delivery with the dynamics of conversation. Prosody, pacing, and emotion shape how claims are received, while dialogue order, repetition, and cross-turn dependencies sustain narratives across speakers and episodes. These dual properties make audio platforms fundamentally different from text or images, revealing why pipelines built for written claims fail when applied to spoken dialogue. We argue that fact-checking must be reframed around these realities, addressing challenges in real-time detection, multimodal fusion, conversational modeling, temporal tracking, and fairness.

\section*{Limitations}

External validity is restricted. Most examples and analyses center on English, long-form talk (podcasts) with a two-speaker structure and reasonably clean production metadata. We do not test generalization to multi-party debates, call-in shows, highly code-switched speech, or low-metadata settings (sparse chapters and show notes), so portability to those regimes remains unverified.

Our pipeline assumptions depend on upstream components that we do not fully characterize. We rely on off-the-shelf ASR, VAD/segmentation, and diarization, but we do not provide a systematic error analysis by error type—e.g., punctuation, proper nouns, code-switch boundaries—or by speaker or acoustic condition, and we do not compare alternative ASR, VAD, or diarization backbones. As a result, the end-to-end robustness of the proposed design under diverse recording conditions remains uncertain.

Misinformation can also be propagated through synthetic or cloned voices, and impersonation via voice cloning is an increasingly serious threat. In this work, however, we treat deepfake and voice-cloning detection as an upstream, separate problem of media integrity and speaker authenticity, and focus instead on the truthfulness of the claims made in spoken dialogue. In our setting, a voice-cloned clip that makes a factually correct statement should still be labeled as \emph{True} at the verification stage, while a separate deepfake detector would be responsible for flagging the audio as synthetic. A full survey of audio deepfake generation and detection is therefore beyond our scope but represents an important complementary line of work to the spoken fact-checking pipeline we analyze here.

Finally, we acknowledge unmeasured risks. We do not quantify misattribution risk from diarization/alignment errors, possible demographic bias (accent, pitch range, speaking style) in prosody-aware features or ASR, or amplification harms from replaying salient misinformation clips. While we discuss safeguards (e.g., timestamped evidence and recoverability checks), the paper does not include bias audits or privacy analyses to demonstrate that these mitigations reduce harm in practice.

\section*{Acknowledgments}
This work was supported in part by U.S. NSF awards \#2114824 and \#2438810. Some experimental results were obtained using computational resources provided by CloudBank, supported through U.S. NAIRR award \#240336.


\bibliography{custom}




\end{document}